\documentclass[runningheads]{llncs}
\usepackage[T1]{fontenc}
\usepackage{graphicx,verbatim}
\usepackage[pagebackref,breaklinks,colorlinks]{hyperref}
\usepackage{multirow}
\usepackage{multicol}
\usepackage{soul}
\usepackage{pifont}
\def\our{PR-ENDO}
\def\ourMLP{diffuseMLP}
\usepackage{amsmath} 
\usepackage{cleveref}
\usepackage{booktabs}

\begin{document}

\title{\our{}: Physically Based Relightable Gaussian Splatting for Endoscopy}
\author{Joanna Kaleta\inst{1,2} \thanks{equal contribution} \and
Weronika Smolak-Dyżewska\inst{3} \textsuperscript{$\star$} \and
Dawid Malarz\inst{4} \and
Diego Dall'Alba\inst{1,5} \and
Przemysław Korzeniowski\inst{1} \and
Przemysław Spurek\inst{4,6}}

\authorrunning{J. Kaleta et al.}
%
\institute{
Sano Centre for Computional Medicine, Poland \and Warsaw University of Technology, Poland
\email{j.kaleta@sanoscience.org}\\
\and
Doctoral School of Exact and Natural Sciences, Jagiellonian University, Poland \and
Faculty of Mathematics and Computer Science, Jagiellonian University, Poland \and
University of Verona, Italy \and
IDEAS Research Institute, Poland
}

\maketitle              
\begin{abstract}

Endoluminal endoscopic procedures are essential for diagnosing colorectal cancer and other severe conditions in the digestive tract, urogenital system, and airways. 3D reconstruction and novel-view synthesis from endoscopic images are promising tools for enhancing diagnosis. Moreover, integrating physiological deformations and interaction with the endoscope enables the development of simulation tools from real video data. However, constrained camera trajectories and view-dependent lighting create artifacts, leading to inaccurate or overfitted reconstructions. We present \our{}, a novel 3D reconstruction framework leveraging the unique property of endoscopic imaging, where a single light source is closely aligned with the camera. Our method separates light effects from tissue properties. \our{} enhances 3D Gaussian Splatting with a physically based relightable model. We boost the traditional light transport formulation with a specialized MLP capturing complex light-related effects while ensuring reduced artifacts and better generalization across novel views. \our{} achieves superior reconstruction quality compared to baseline methods on both public and in-house datasets. Unlike existing approaches, \our{} enables tissue modifications while preserving a physically accurate response to light, making it closer to real-world clinical use. Repository: \url{https://github.com/SanoScience/PR-ENDO}.

\end{abstract}

\begin{figure*}[t]
    \centering

    \includegraphics[width=0.99\textwidth, trim=0 0 0 0, clip]{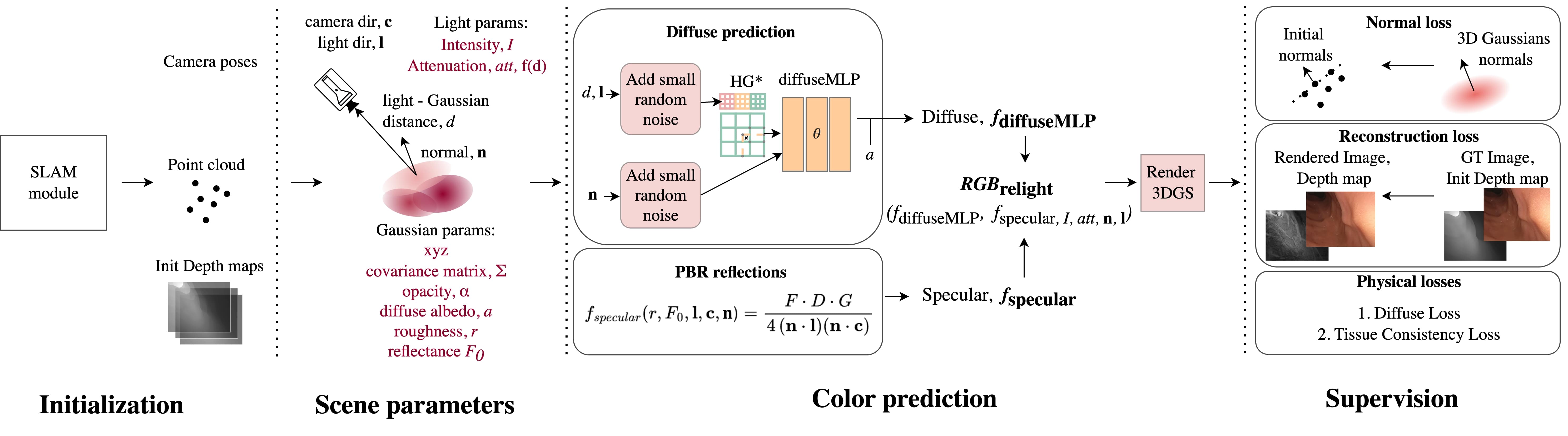}
    
    \caption{\textbf{Overview of \our{}} pipeline: (1) \textbf{Initialization}: SLAM module generates camera poses and point cloud and optionally depth maps. (2) \textbf{Scene Parameters}: Each Gaussian \( i \) is defined by position, covariance \( \Sigma_{i} \), opacity \( \alpha_{i} \), albedo \( a_{i} \), roughness \( r_{i} \), and reflectivity \( F_{0,i} \). Light parameters include direction \( \mathbf{l} \), intensity \(I\) and attenuation coefficients for \( att \). (3) \textbf{Prediction}: \(\ourMLP{} \) predicts the diffuse component, while the PBR model calculates specular reflection. The final color \( \mathrm{RGB}_{\mathrm{relight}} \) combines diffuse and specular terms. (4) \textbf{Supervision}: Normal, reconstruction and physical losses are applied during training.
}
    \label{fig:architecture}
\end{figure*}
\section{Introduction}
\label{sec:intro}
 
Endoluminal Endoscopy (EE) is a minimally invasive technique widely used for diagnosing and treating complex conditions, including tumors, within luminal structures such as the airways, digestive tract, and urogenital system \cite{pore2023autonomous}. Despite its clinical significance, EE remains technically challenging due to the complexity of maneuvering a flexible endoscope in constrained environments. These difficulties can lead to incomplete inspections and risks to patient safety \cite{pore2023autonomous}. 

High-fidelity, interactive 3D reconstructions of luminal structures offer a powerful solution by allowing clinicians to revisit previously examined areas, synthesize novel views, and improve diagnostic accuracy \cite{ma2021rnnslam}. Additionally, accurate lumen reconstructions facilitate realistic, patient-specific training environments \cite{dlInEndoReview} and provide valuable data for developing autonomous robotic platforms \cite{roboBiopsy}, \cite{pore2023autonomous}. 
Recent advances in Neural Radiance Fields (NeRF) \cite{mildenhall2021nerf} and Neural Implicit Surfaces (NeuS) \cite{wang2021neus} have been applied to EE, particularly in colonoscopy, for novel view synthesis \cite{shi2023colonnerf}, \cite{batlle2023lightneus}, \cite{psychogyios2023REIM-NeRF}. However, these methods are computationally expensive, require long training and rendering times and are hindered by non-explicit, difficult to edit representation. 3D Gaussian Splatting (3DGS) \cite{kerbl20233dgs} overcomes these challenges by modeling scene with Gaussian primitives and efficient rasterization pipeline. Several studies have explored 3DGS for endoscopic reconstruction \cite{endosurf}, \cite{wang2022neural}, \cite{Li_EndoSparse_MICCAI2024}, \cite{Hua_Endo4DGS_MICCAI2024}. Especially for EE, \cite{bonilla2024gaussianpancakes} enhances 3DGS with geometric and depth regularization, while \cite{wang2024endogslam} jointly estimates camera poses and reconstruction. However, none of these works address the challenges posed by endoscopic lighting. As a result, illumination effects are typically baked into textures or mimicked by "floaters", leading to inaccurate reconstructions.

Inverse rendering techniques have been developed to enhance 3DGS with separation of lighting effects from object material properties \cite{jiang2024gaussianshader}, \cite{ye2024gsdr}, \cite{chen2024gi-gs}, \cite{gao2023relightable}, \cite{liang2024gs-ir}, \cite{sss_gs},
\cite{kaleta2025lumigauss}
. Most approaches represent materials with the use of Bidirectional Reflectance Distribution Function (BRDF). Malarz et al. \cite{malarz2024gaussiansplattingnerfbasedcolor} model camera-dependent effects but it lacks physical interpretation. None of these works specifically consider EE challenges, such as very constrained camera view with a strong view dependent illumination. Moreover, none have explored a physically correct editable model, a crucial feature for medical simulation.

In this work, we propose \our{}, a 3D reconstruction framework that leverages the co-location of the endoscopy camera and the light source. Built on 3DGS, our method offers fast training and rendering while separating illumination effects from tissue properties. \our{} extends traditional light transport modeling with a specialized MLP that captures complex lighting interactions, reduces artifacts, and improves generalization across novel views. Our approach enables accurate re-illumination of luminal structures, including cases where tissue properties or geometry are arbitrarily modified, making it highly applicable to real-world clinical use. We evaluate our framework on two datasets: one public \cite{bobrow2023c3vd} and another introduced in this work to provide a more diverse and challenging set of test camera rotations.   

Our contributions can be summarized as follows:  
(1) A 3D relightable reconstruction framework that models interactions between tissue and endoscopic light source in a physically correct way.  
(2) The enhancement of traditional light transfer formulation with  \textit{diffuseMLP}, to capture complex lighting effects and provide generalizable novel view synthesis and superior reconstruction quality.
(3) A demonstration of our method’s capability to handle view-dependent lighting effects not only for novel views, but also for novel anatomical movements.

\begin{figure*}[tb]
    \centering
    \includegraphics[width=0.75\textwidth, trim=0 0 0 0, clip]{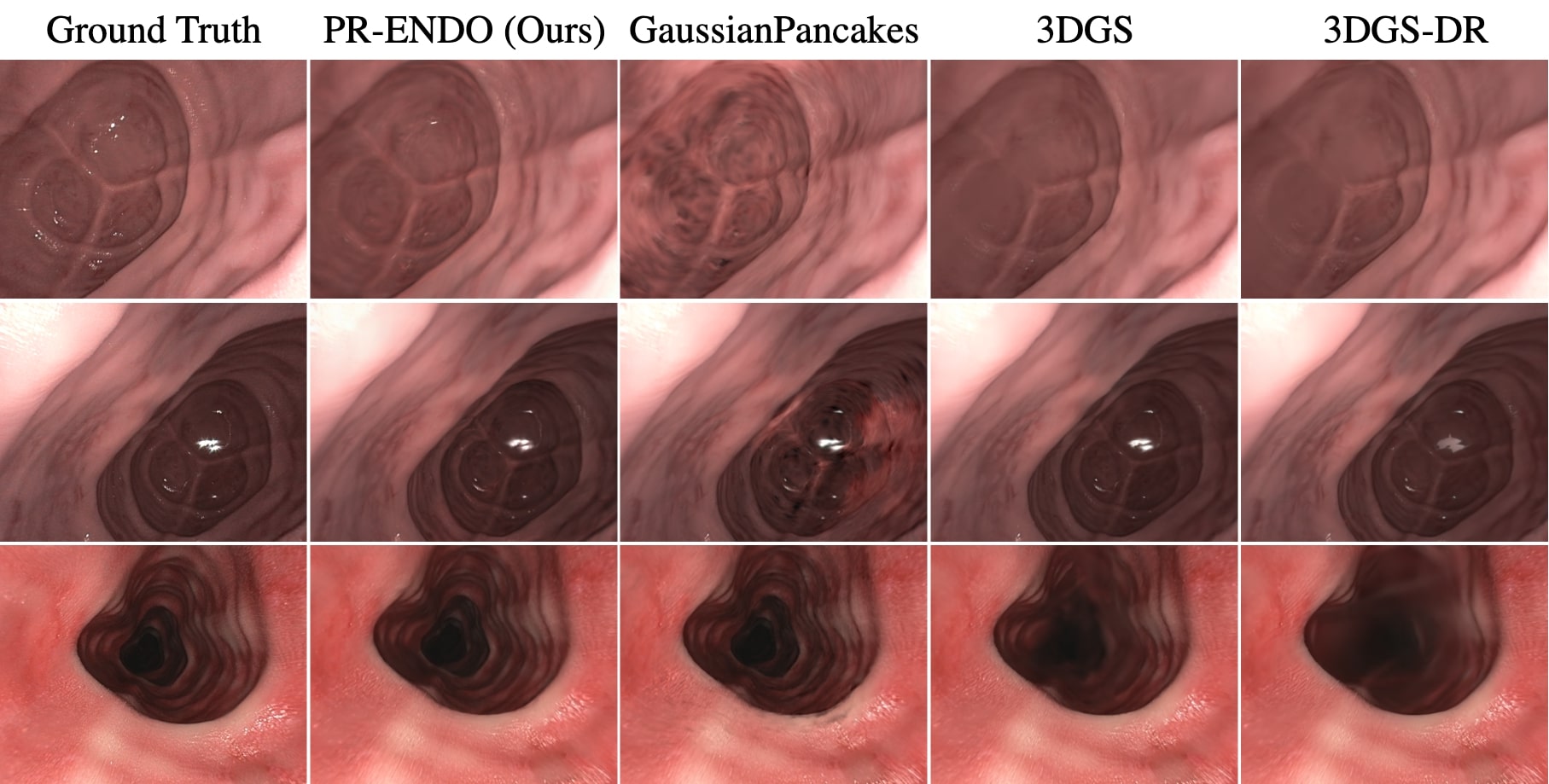}
    
    \caption{\textbf{Qualitative comparison} of \our{} and baseline methods. Our method achieves strong reconstruction quality and produces fewer artifacts than other works. We strongly encourage to view supplementary videos, as some artifacts, e.g. "floaters" imitating light-dependent effects, are only clearly visible there}.
    \label{fig:c3vd}
\end{figure*}

\section{Method}
\label{sec:methods}

This section describes our model. First, we present the preliminaries on vanilla 3DGS algorithm \cite{kerbl20233dgs} and physically based rendering. Then, we present \our{}, which combines 3DGS with a relighting model, dedicated to EE data.  

\textbf{3DGS} \cite{kerbl20233dgs} reconstructs a 3D scene from a set of images. It represents the scene as a collection of 3D Gaussian primitives, each characterized by a mean position, a covariance matrix encoding spatial uncertainty, an opacity term, and a view-dependent color. The covariance matrix is parameterized using a scaling matrix \( S_i \) and a rotation matrix \( R_i \), forming:  
\begin{equation}
    \Sigma_i = R_i S_i S_i^T R_i^T.
\end{equation}

Rendering projects 3D Gaussians onto a 2D image plane by transforming their covariance matrix in screen space:
\begin{equation}
    \Sigma_i' = J W \Sigma_i W^T J^T,
\end{equation}
where \( W \) is the viewing transformation, and \( J \) accounts for perspective distortion. The final image is synthesized via differentiable alpha blending, with each Gaussian contributing based on opacity and position. The accumulated pixel color is:
\begin{equation}
    C = \textstyle\sum_{i \in N} T_i \alpha_i c_i, \quad T_i = \prod_{j=1}^{i-1} (1 - \alpha_j),
\end{equation}
where \( \alpha_i \) is per-Gaussian opacity, modulated by its transformed covariance. For exact derivation and further details on the formulation and rendering process, we refer to \cite{kerbl20233dgs}.

\begin{figure}[t]
    \centering

    \includegraphics[width=0.99\textwidth, trim=0 0 0 0, clip]{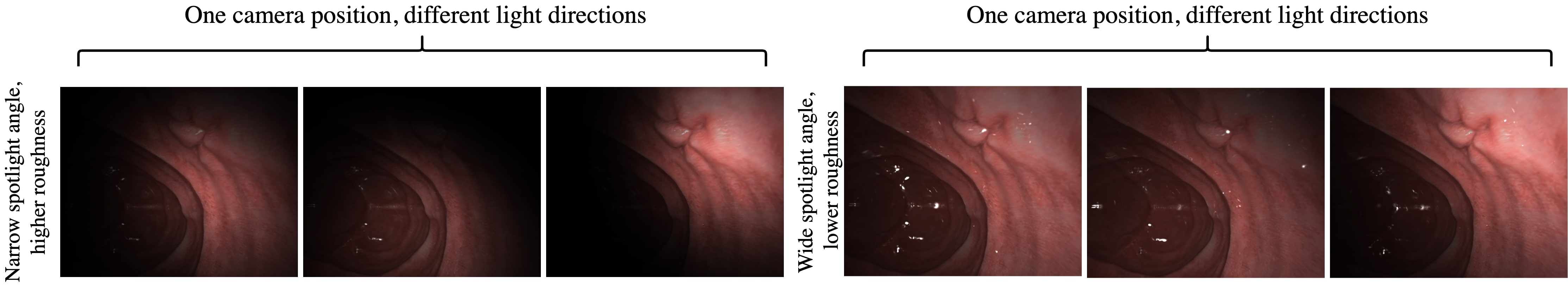}
    
    \caption{\textbf{Qualitative evaluation of light and tissue properties separation.} \our{} enables effective novel camera positioning, light parameters adjustments, and tissue properties modifications - capabilities not possible in vanilla 3DGS \cite{kerbl20233dgs}.}
    \label{fig:lighting_eval}
\end{figure}

\textbf{Physically-Based Rendering (PBR).}  
The basics for light transfer and rendering equation can be found in \cite{veach1995bidirectional}. To enable physically-based relighting of 3D Gaussians  we associate each Gaussian with BRDF properties.
Surface reflectance follows the \textbf{BRDF} with Cook-Torrance microfacet model \cite{cooktorr}:
\begin{equation}
\text{BRDF} = k_d f_{\text{diffuse}} + k_s f_{\text{specular}}
\end{equation}
where \( k_d \) and \( k_s \) controls the diffuse and specular contribution respectively and sum to 1.0. \( f_{\text{diffuse}} \) and \( f_{\text{specular}} \) represent the diffuse and specular reflections.

The diffuse term \( f_{\text{diffuse}} \) is proportional to albedo. The specular term \( f_{\text{specular}} \) depends on light and view directions and is formulated as:
\begin{equation}
    f_{\text{specular}} = \frac{D(h, r) \cdot F(\mathbf{c}, h, F_0) \cdot G(\mathbf{l}, \mathbf{c}, h, r)}
    {4 \cdot(\mathbf{n} \cdot \mathbf{l}) \cdot (\mathbf{n} \cdot \mathbf{c})},
    \label{eq:specular_term}
\end{equation}
where \( r \) is the roughness, \( F_0 \) is the base reflectance, \( \mathbf{l} \) is the light direction, \( \mathbf{c} \) is the view (camera) direction, \( h = (\mathbf{l} + \mathbf{c})/2 \) is the half-vector and \( \mathbf{n} \) is the surface normal. The terms \( D \), \( F \), and \( G \) correspond to the microfacet distribution function, Fresnel effect, and geometry term, respectively. We employ the Trowbridge-Reitz GGX distribution for \( D \), Schlick's approximation for \( F \), and the Schlick-Beckmann model for \( G \). For further details, we refer to \cite{cooktorr}, \cite{disney2012pbr}.

\begin{figure*}[t]
    \centering

    \includegraphics[width=0.95\textwidth, trim=0 0 0 0, clip]{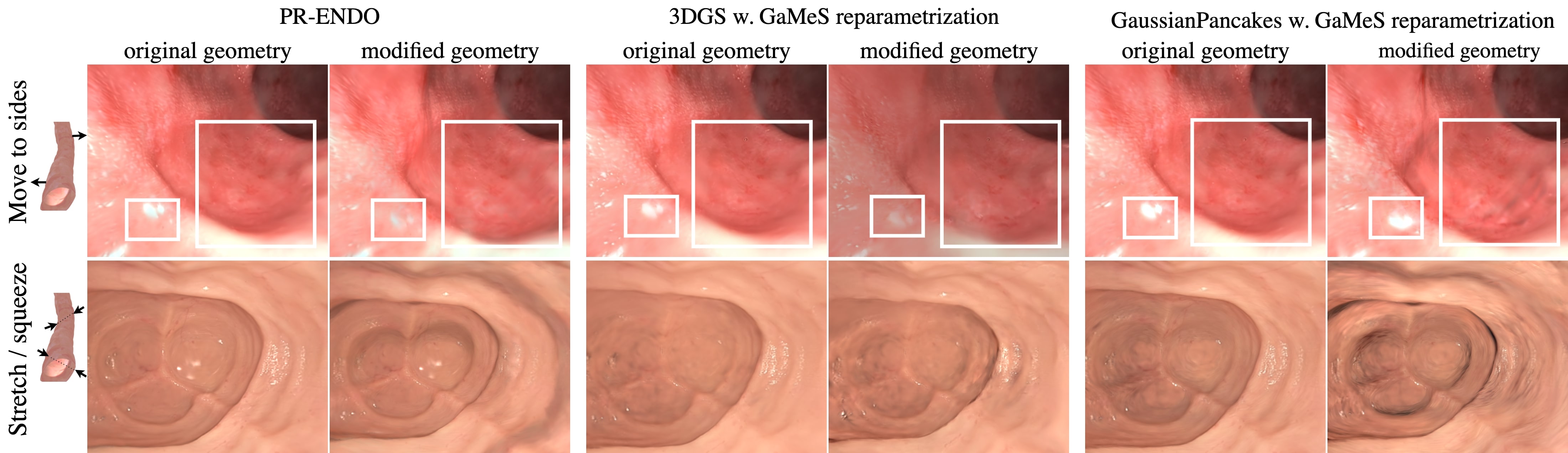}
    
    \caption{\textbf{Anatomy modification using \cite{waczynska2024games} and cage-based simulations.} \our{} ensures accurate tissue responses to light, unlike 3DGS \cite{kerbl20233dgs} and GaussianPancakes \cite{bonilla2024gaussianpancakes}, which bake light-dependent effects into textures. \our{} significantly reduces artifacts such as floaters and decolorized body fragments.}
    \label{fig:body_move}
\end{figure*}

\textbf{\our{}.} Our method, \our{}, introduces a 3DGS physically-based relighting model tailored for endoscopic scenes, capturing both diffuse and specular interactions between light and surfaces.   

We show the overview of our method in \cref{fig:architecture}. Like in most GS-based works, including \cite{bonilla2024gaussianpancakes}, a SLAM module is used for initialization and camera poses. \our{} separates the scene representation into a light model and set of flatten \cite{waczynska2024games} relightable Gaussians, with additional parameters: base color \(a\) , roughness \(r\) and base reflectivity \(F_0\). We use a dedicated light model, which assumes that the point light source and camera are co-located. The precise positioning of the light source, slightly shifted from the camera center, can be jointly optimized with other light parameters.  

Given the limited camera trajectory and complexity of lumen geometry and light behavior, for diffuse irradiance we enhance the traditional BRDF \cite{cooktorr} with \ourMLP{} \( \theta \).  The \ourMLP{} takes as input the Gaussian normal \( \mathbf{n_i} \) and light position parameters (\( \mathbf{l_i} \), \( d_i \)) relative to the \textit{i}-th Gaussian. To enhance robustness across lighting and viewpoint variations, we introduce small random noise to the inputs, encouraging consistent diffuse predictions. The output models the diffuse component of the \textit{i}-th Gaussian’s:
\begin{equation}
f_{\text{diffuse}MLP,i} = \theta(\mathbf{l_i}, d_i, \mathbf{n_i}) \cdot \frac{a_{i}}{\pi}.
\end{equation}
This approach improves generalization to unseen rotations and enables additional lighting effects beyond the coefficient-based model.

The specular component \(f_{specular, i}\) for the \textit{i}-th Gaussian is computed using the BRDF model as in \cref{eq:specular_term}. In our method we clamp $F_0$ values to 0.03, which represent realistic values of the colon refractive index \cite{refractive}.

The final relit color, \( RGB_{\text{relight}, i} \), is obtained by combining the diffuse and specular components, scaled by light intensity \( I \), distance-based attenuation \( \text{att}(d_i) \), and dot product between light direction and Gaussian normal:
\noindent
\begin{equation}
RGB_{\mathrm{relight}, i} = 
I \cdot att(d_{i}) \cdot  
\left( f_{\text{diffuseMLP}, i} \cdot (1 - F_{i}) + 
f_{\text{specular}, i} \right) \cdot (\boldsymbol{\textbf{n}}_i \cdot \boldsymbol{\textbf{l}}_i)
\end{equation}

To enhance the effect of light position, we optionally encode the light position using HashGrid (HG), a multi-resolution hash table, as an additional input to \ourMLP{}. This improves the capture of view-dependent effects along the training trajectory.   

\textbf{Optimization.} In our data-driven PBR model, all material and light properties are optimized via gradient descent. Our total loss function \( \mathcal{L} \) incorporates multiple terms from prior work to ensure fidelity, depth alignment, and structural consistency. Specifically, we adopt the image reconstruction loss \( \mathcal{L}_{\text{RGB}} \) and SSIM loss \( \mathcal{L}_{\text{D-SSIM}} \) for perceptual quality, depth loss \( \mathcal{L}_{\text{Depth}} \) from \cite{bonilla2024gaussianpancakes} to handle minimal depth variation, and normal loss \( \mathcal{L}_{\text{Norm}} \) from \cite{bonilla2024gaussianpancakes} to enforce normal consistency.  We introduce two novel losses:  

\textbf{Diffuse Loss}: Ensures consistency between MLP-based irradiance and irradiance computed for point light using classic light transport equation:  
  \begin{equation}
  \mathcal{L}_{\text{Diffuse}} = \textstyle\sum_{i=1}^N \left(f_{\text{diffuseMLP}, i} - f_{\text{diffuse,ClassicEq}, i}\right)^2
  \end{equation}

\textbf{Tissue Consistency Loss}: Regularizes albedo, roughness, and base reflectance across Gaussians:  
  \begin{equation}
  \mathcal{L}_{\text{Tissue}} = \textstyle\sum_{i=1}^N (a_{i} - a_{\text{mean}})^2 + (r_{i} - r_{\text{mean}})^2 + (F_{0,i} - F_{0,\text{mean}})^2
  \end{equation}

The final objective function is:  
\begin{equation}
   \mathcal{L} = \mathcal{L}_{\text{RGB}} + \mathcal{L}_{\text{D-SSIM}} +\mathcal{L}_{\text{Depth}} +
   \mathcal{L}_{\text{Norm}} +\mathcal{L}_{\text{Diffuse}} + \mathcal{L}_{\text{Tissue}} 
\end{equation}  

\begin{figure*}[t]
    \centering

    \includegraphics[width=0.75\textwidth, trim=0 0 0 0, clip]{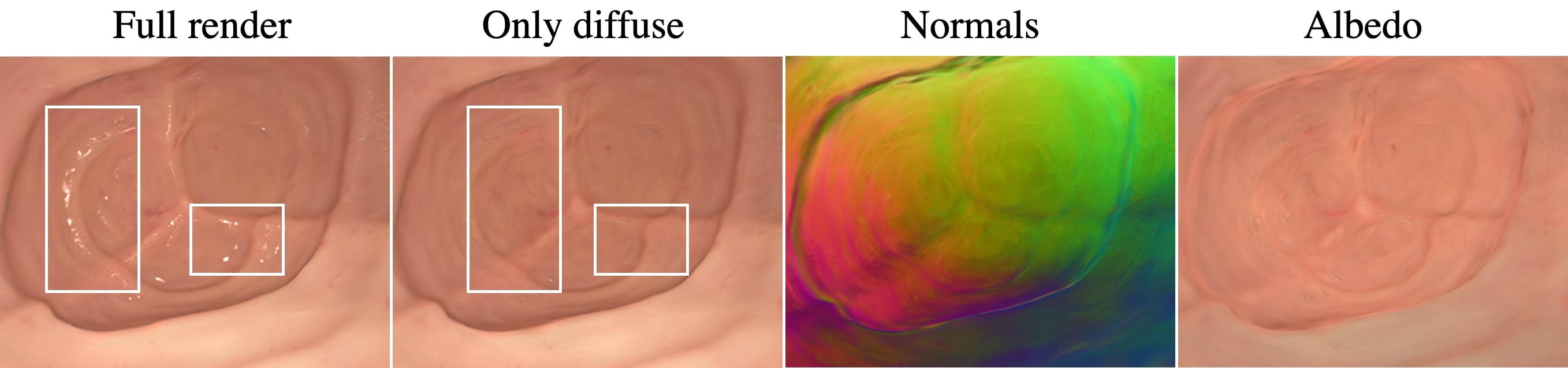}
    
    \caption{\textbf{Decomposition.} Our physically motivated model decomposes the render into specular, diffuse and albedo views.}
    \label{fig:decompose}
\end{figure*}

\begin{table*}[b!]
    \centering
    \fontsize{8pt}{9pt}\selectfont 
    \caption{Performance comparison on C3VD and RotateColon datasets. \our{} achieves superior or comparable performance to state-of-the-art Gaussian Splatting methods for both endoscopy-specific and relighting tasks. *While GaussianShader performs well in terms of metrics, it fails to accurately reconstruct scene geometry. }
    \setlength{\tabcolsep}{3pt} 
    \renewcommand{\arraystretch}{0.9} 
    \begin{tabular}{@{}l ccc ccc@{}}
        \toprule
        \multirow{2}{*}{\textbf{Model}} & \multicolumn{3}{c}{\textbf{C3VD}} & \multicolumn{3}{c}{\textbf{RotateColon}} \\
        \cmidrule(lr){2-4} \cmidrule(lr){5-7}
         & \textbf{PSNR} $\uparrow$ & \textbf{SSIM} $\uparrow$ & \textbf{LPIPS} $\downarrow$ & \textbf{PSNR} $\uparrow$ & \textbf{SSIM} $\uparrow$ & \textbf{LPIPS} $\downarrow$ \\
        \midrule
        3DGS \cite{kerbl20233dgs} & 33.90 & 0.89 & 0.28 & 20.29 & 0.82 & \bf 0.25 \\
        EndoGSLAM \cite{wang2024endogslam} & 22.16 & 0.77 & \bf 0.22 & - & - & - \\ 
        GaussianPancakes \cite{bonilla2024gaussianpancakes} & 33.12 & 0.89 & 0.30 & 20.10 & 0.88 & 0.27 \\ 
        GaussianShader \cite{jiang2024gaussianshader} * & 29.82 & 0.86 & 0.40 & 21.25 & 0.87 & 0.38 \\                
        3DGS-DR \cite{ye2024gsdr} & 33.77 & 0.89 & 0.31 & 21.49 & \bf 0.89 & 0.28 \\      
        \our{} (ours) & \bf 34.24 & \bf 0.90 & 0.29 & \bf 21.90 & 0.88 & 0.27 \\                       
        \bottomrule
    \end{tabular}
    \label{tab:performance_comparison}
\end{table*}

\section{Experiments and Results}
\label{sec:exp}

\textbf{Datasets:} We evaluate our method on the Colonoscopy 3D Video Dataset (C3VD) \cite{bobrow2023c3vd}, following preprocessing from EndoGSLAM \cite{wang2024endogslam}. RGB images for training are pre-undistorted with a resolution of 675×540. Additionally, C3VD provides GT depth maps. The test set for each scene contains every 8-th frame.

Alongside C3VD, we introduce RotateColon, a dataset for novel view synthesis under extended rotations, generated using our in-house simulator. Unlike traditional methods sampling every n-th frame from training sequence, it features intense unseen rotations for stricter generalization testing. Training images have a 640×640 resolution, ground truth camera poses are available.

\begin{figure}[t]
    \centering

    \includegraphics[width=0.9\textwidth, trim=0 0 0 0, clip]{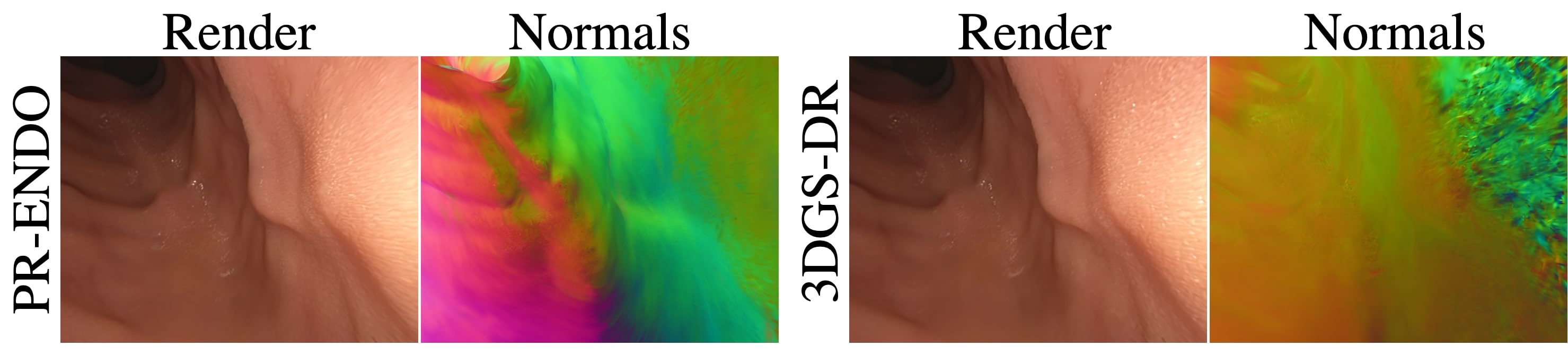}
    
    \caption{\textbf{Normal comparison between \our{} and 3DGS-DR \cite{ye2024gsdr}.} Although both methods produce plausible renderings, thanks to building on \cite{bonilla2024gaussianpancakes} our method offers a superior normals reconstruction.}
    \label{fig:normal_comparison}
\end{figure}

\begin{figure}[b]
    \centering

    \includegraphics[width=0.75\textwidth, trim=0 0 0 0, clip]{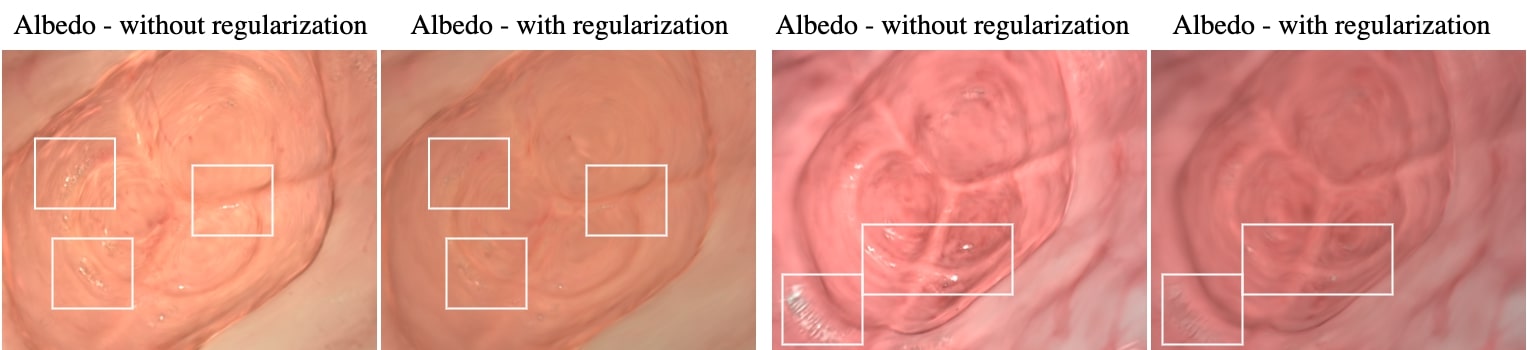}
    
    \caption{\textbf{Ablation study on albedo regularization.} Tissue regularization \(\mathcal{L}_{\text{Tissue}}\) prevents reflections and lighting effects from becoming part of the base color.}
    \label{fig:albedo}
\end{figure}

\textbf{Metrics:} We measure rendering quality using widely approved set of metrics: peak signal-to-noise ratio (PSNR), structural similarity index (SSIM), and learned perceptual image patch similarity (LPIPS).  

\textbf{Initialization and Training Details: }  
 We expect our method to demonstrate strong performance with SLAM modules designed for lumen geometry reconstruction, such as RNNSLAM \cite{ma2021rnnslam}, used in Gaussian Pancakes \cite{bonilla2024gaussianpancakes}. However, since \cite{ma2021rnnslam} was not open-sourced at the time of submission, we substituted it with EndoGSLAM \cite{wang2024endogslam}, which similarly estimates camera poses and generates an initial point cloud. For both our method and baselines, we use EndoGSLAM for initialization and camera poses for C3VD dataset. All training configurations are provided in our repository.
 
\textbf{Reconstruction:}  
As shown in \cref{tab:performance_comparison}, \our{} achieves strong reconstruction performance, surpassing or matching 3DGS \cite{kerbl20233dgs}, EndoGSLAM \cite{wang2024endogslam}, GaussianPancakes \cite{bonilla2024gaussianpancakes}, and relighting methods such as GaussianShader \cite{jiang2024gaussianshader} and 3DGS-DR \cite{ye2024gsdr}. Notably, quantitative results demonstrate better generalization for RotateColon under challenging rotations. Fig. \ref{fig:c3vd} highlights reduced artifacts from light-dependent effects compared to GS-based methods that do not decompose light, and shows improved reflections compared to 3DGS-DR.

\textbf{Intrinsics Decomposition:}  
\our{} accurately decomposes diffuse, specular reflections, and albedo (\cref{fig:decompose}) while, thanks to additional regularizations, producing more plausible normals than existing methods (\cref{fig:normal_comparison}).   

\textbf{ Novel Light Position and Anatomy Movement:} 
By modeling tissue-light interactions, \our{} enables dynamic light adjustments in novel views, including previously unseen rotations, setting it apart from other methods. It effectively separates light from tissue properties, as shown in \cref{fig:lighting_eval}. Proper response to light is also achieved for cage-based geometry modifications (e.g., using reparametrization from \cite{waczynska2024games}), as demonstrated in \cref{fig:body_move}, making it suitable for physically accurate colon motion simulations. \textbf{We showcase high rendering quality and artifact reduction in the supplementary videos.}

\textbf{Ablations:}  
We evaluate the impact of key components and losses in \cref{tab:ablation}. \( \mathcal{L}_\text{Tissue} \) is essential for proper component separation, as it prevents the albedo from absorbing reflections and lighting effects (\cref{fig:albedo}). While  HG improves reconstruction along training trajectory (\cref{tab:hg}), it may reduce generalization in challenging rotations, making its inclusion user-dependent.

\textbf{Limitations:}  
\our{} offers fast optimization and rendering. However, training time is $\sim 2\times$ longer than for vanilla 3DGS. Our model improves generalization to novel viewpoints and geometries compared to existing methods but still faces challenges with with very distant light sources, extreme camera rotations and extreme body deformations. For highly textured tissues, a more complex \(\mathcal{L}_{Tissue}\) may be required.

\begin{table*}[t]
    \centering
    \begin{minipage}{0.45\textwidth}
        \setlength{\tabcolsep}{2pt} 
        \renewcommand{\arraystretch}{0.9} 
        \fontsize{8pt}{9pt}\selectfont 
        \centering
        \caption{Study on HashGrid (HG) influence in the considered datasets.} 
        \begin{tabular}{ccccc}
            \toprule
            \textbf{Dataset} & \textbf{HG} & \textbf{PSNR} $\uparrow$ & \textbf{SSIM} $\uparrow$ & \textbf{LPIPS} $\downarrow$ \\  
            \midrule
            \multirow{2}{*}{C3VD} &  \ding{55}  & 33.65 & 0.89 & 0.30 \\  
                                  & \ding{51}  & 34.24 & 0.90 & 0.29 \\                              
            \midrule                      
            {Rotate} &  \ding{55}  & 21.90 & 0.88 & 0.27 \\  
            {Colon} & \ding{51}  & 19.95 & 0.86 & 0.27 \\                                                      
            \bottomrule
        \end{tabular}
        \label{tab:hg}
    \end{minipage}
    \hfill
    \begin{minipage}{0.45\textwidth}
        \setlength{\tabcolsep}{2pt} 
        \renewcommand{\arraystretch}{0.9} 
        \fontsize{8pt}{9pt}\selectfont 
        \centering
        \caption{Ablation study on C3VD set considering a model without HG.} 
        \begin{tabular}{lccc}
            \toprule
            \textbf{\our{}} & \textbf{PSNR} $\uparrow$ & 
            \textbf{SSIM} $\uparrow$ & \textbf{LPIPS} $\downarrow$ \\  
            without & & & \\
            \midrule
            - & 33.65 & 0.89 & 0.30 \\  
            \ourMLP{}  & 32.69 & 0.89 & 0.31 \\  
            \(L_\text{Tissue}\)  & 33.71 & 0.89 & 0.29 \\   
            \(L_\text{Diffuse}\)  & 33.43 & 0.89 & 0.30 \\  
            \bottomrule
        \end{tabular}
        \label{tab:ablation}
    \end{minipage}
\end{table*}

\section{Conclusions}
\label{sec:conclusion}
\our{} is a model specifically designed for 3D reconstruction of endoscopic videos. It leverages the unique property of endoscopic imaging, where a single light source is closely aligned with the camera, enabling a physically accurate separation of light effects from tissue properties. This allows for realistic reflections, improved generalization, and the ability to modify tissue properties and geometry while maintaining physically consistent lighting. As a result, PR-ENDO produces high-quality novel views, even at challenging angles. Qualitative and quantitative comparisons show that PR-ENDO surpasses previous state-of-the-art methods.

\textbf{Acknowledgments} The work of W. Smolak-Dyżewska, D. Malarz and P. Spurek was supported by the project \textit{Effective Rendering of 3D Objects Using Gaussian Splatting in an Augmented Reality Environment} (FENG.02.02-IP.05-0114/23), carried out under the First Team programme of the Foundation for Polish Science and co-financed by the European Union through the European Funds for Smart Economy 2021–2027 (FENG). The work of J. Kaleta was supported by National Science Centre, Poland (grant no. 2022/47/O/ST6/01407). This research was supported by the European Union’s Horizon 2020 programme under grant agreement Sano No. 857533, and by the Sano project within the International Research Agendas programme of the Foundation for Polish Science, co-financed by the European Regional Development Fund. This research was also supported by the Polish Ministry of Science and Higher Education under the “Support for the activity of Centers of Excellence established in Poland under Horizon 2020” programme (contract no. MEiN/2023/DIR/3796).

\textbf{{\discintname}} The authors have no competing interests to declare that are relevant to the content of this article. 

\bibliographystyle{splncs04}

\clearpage
\section{Supplemetary Material}

\subsection{Detailed Endoscopic Setup}

\begin{figure}[h!]
    \centering
    \includegraphics[width=0.99\textwidth, trim=0 0 0 0, clip]{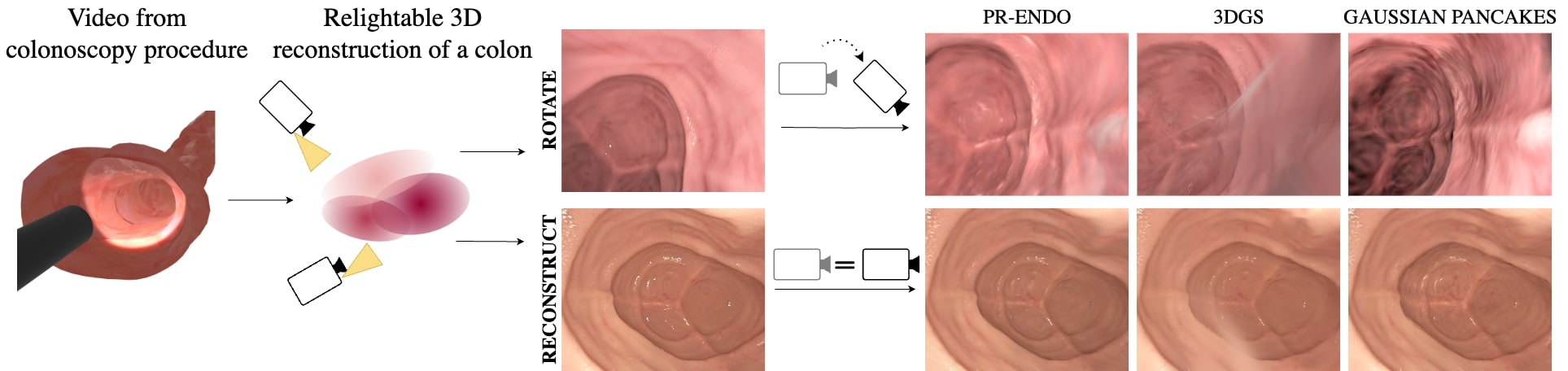}
    \caption{PR-ENDO adapts classical Gaussian Splatting for endoscopy environments. In contrast to 3DGS, we use a physical light model that incorporates the connection between the camera and the light source. Moreover, we reduce artifacts during novel view generation—a common issue due to limited camera trajectory during training. Finally, our model generates fewer artifacts than other state-of-the-art methods and responds effectively to novel lighting conditions when the viewpoint is changed or the geometry is modified.}
    \label{fig:teaser}
\end{figure}

\begin{figure}[h!]
    \centering
    \includegraphics[width=0.40\textwidth, trim=0 0 0 0, clip]{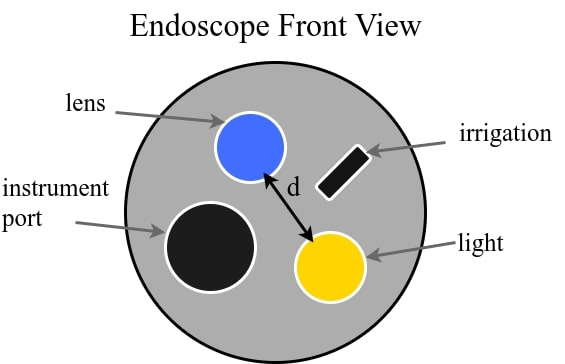}
    \caption{\textbf{Detailed colonoscope architecture.} The light and camera sources are colocated, with a small distance between them, denoted as \(d\). We \textbf{optimize \(d\)} alongside other light parameters.}
    \label{fig:detailed_scheme}
\end{figure}

The setup overview is shown in \cref{fig:teaser}. The exact positioning of the light source relative to the lens is illustrated in \cref{fig:detailed_scheme}. Our setup supports seamless optimization of the spotlight angle for the light source. While this feature was not utilized in our experiments due to the characteristics of our datasets, it could be jointly optimized alongside other light parameters. Additionally, experiments involving adjustments to the spotlight angle during the inference stage are shown in \cref{sec:relightning_additional_experiments}.

\subsection{Additional Experiments on Relighting}
\label{sec:relightning_additional_experiments}
\begin{figure*}[t!]
    \centering
    \includegraphics[width=0.80\textwidth, trim=0 0 0 0, clip]{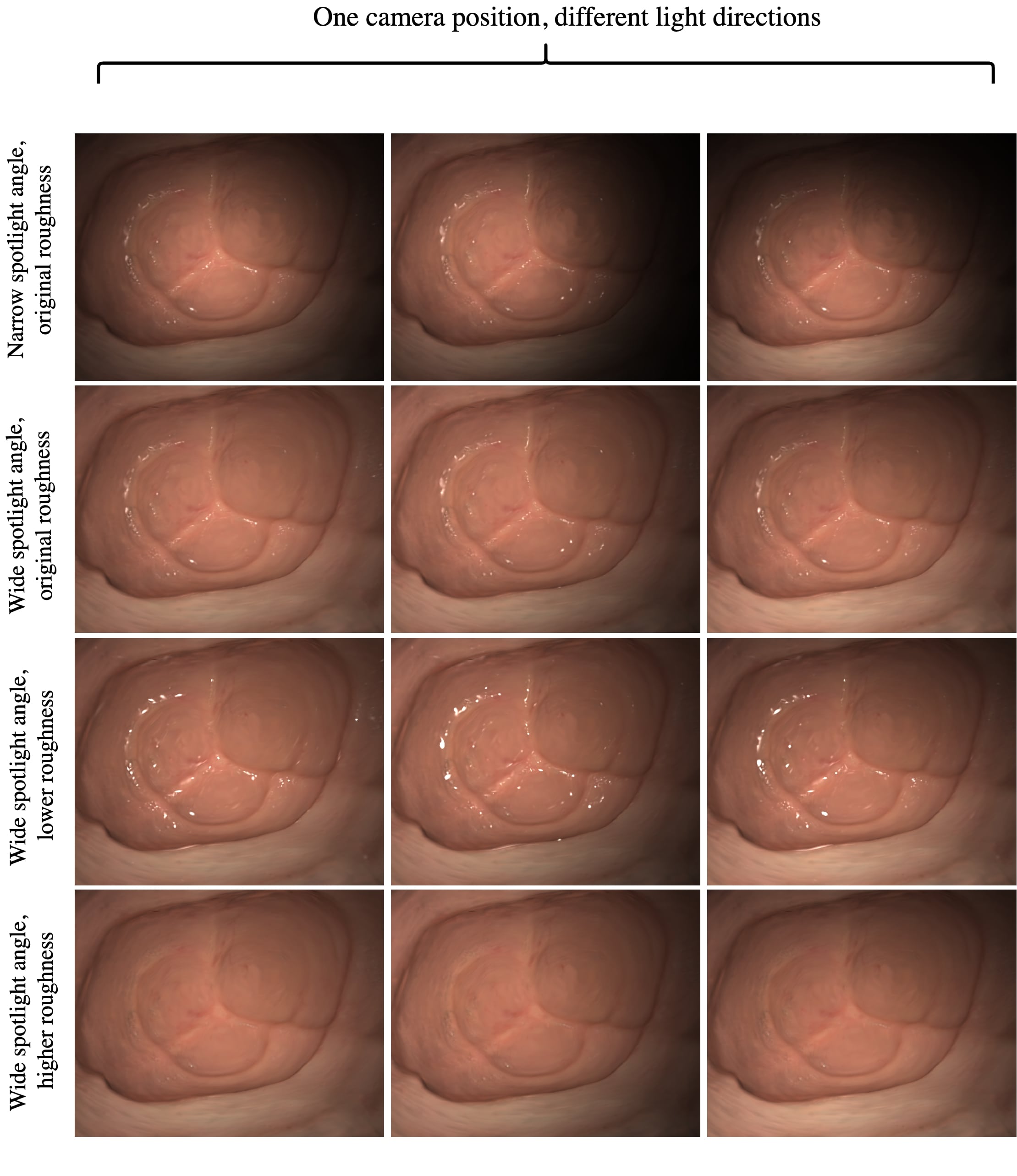}
    \caption{\textbf{Additional relighting results for Scene cecum t1b.} Light direction, spotlight angle and Gaussian roughness are manipulated to explore their effect on specularity and surface appearance.}
    \label{fig:relightining1}
\end{figure*}

\begin{figure*}[t!]
    \centering
    \includegraphics[width=0.80\textwidth, trim=0 0 0 0, clip]{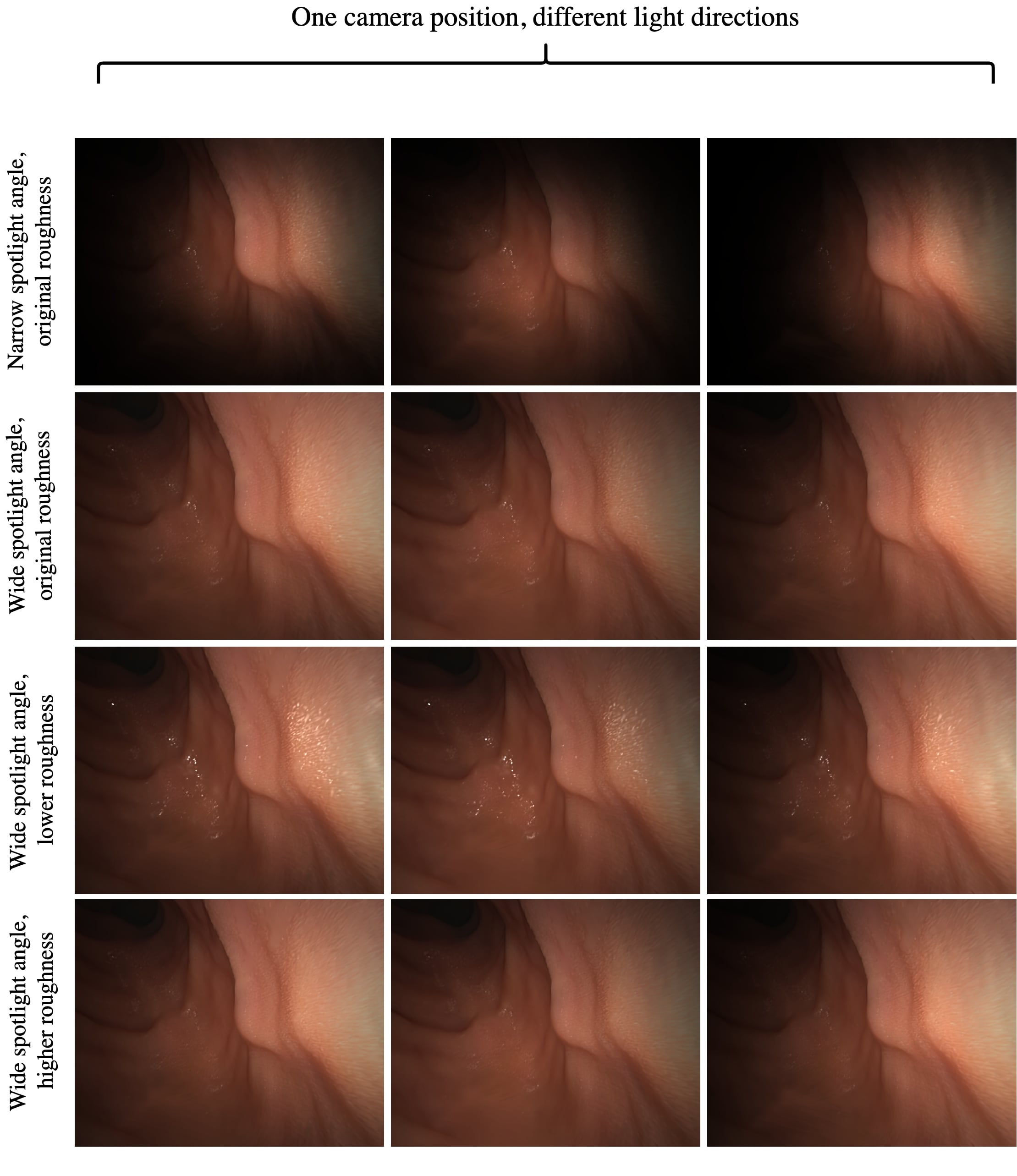}
    \caption{\textbf{Additional relighting results for Scene sigmoid t3b.} Light direction, spotlight angle and Gaussian roughness are manipulated to explore their effect on specularity and surface appearance.}
    \label{fig:relightining2}
\end{figure*}

\begin{figure*}[t!]
    \centering
    \includegraphics[width=0.80\textwidth, trim=0 0 0 0, clip]{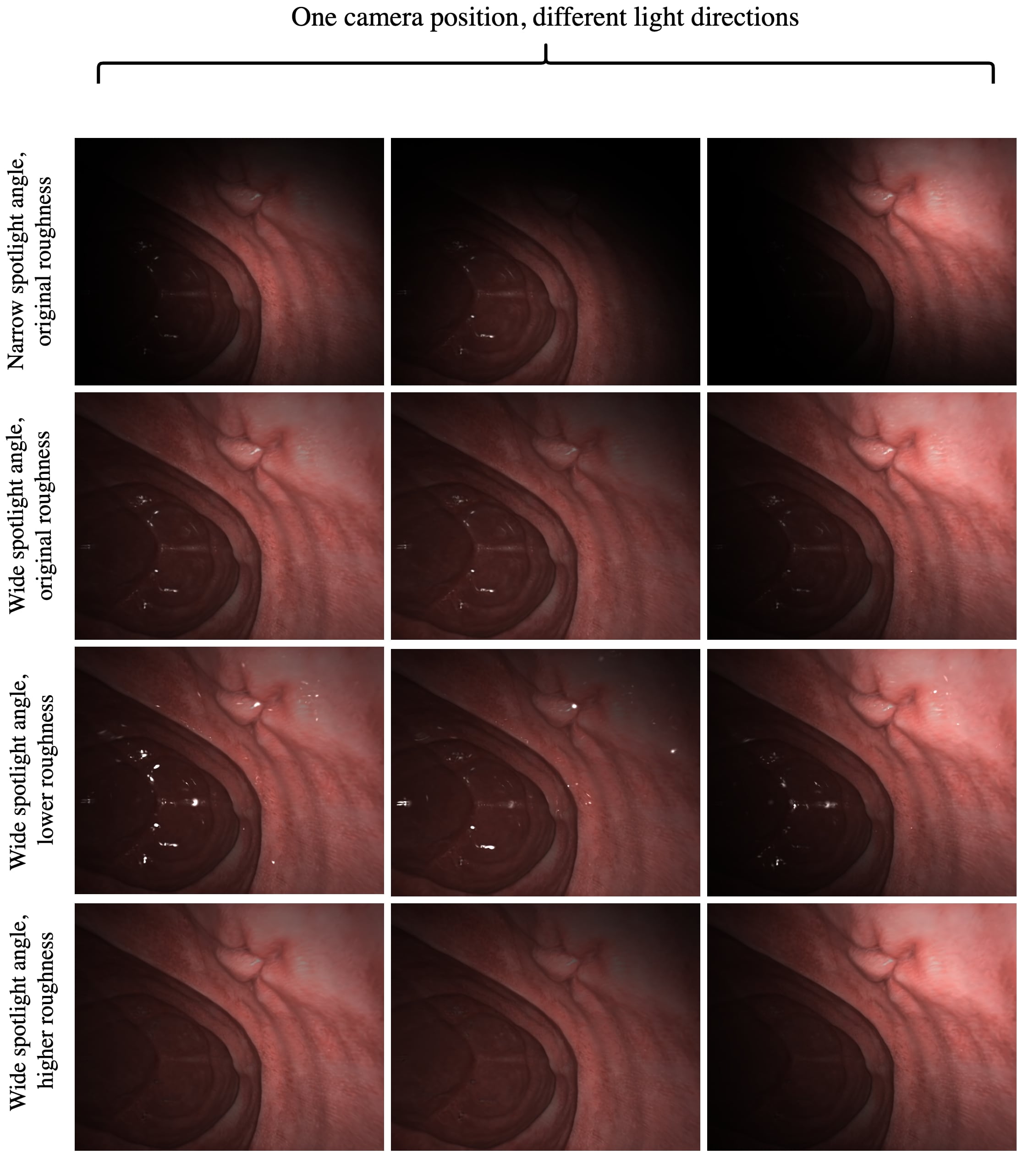}
    \caption{\textbf{Additional relighting results for Scene cecum t3b.} Light direction, spotlight angle and Gaussian roughness are manipulated to explore their effect on specularity and surface appearance.}
    \label{fig:relightining3}
\end{figure*}

We demonstrate the manipulation of the light source and physical parameters in \cref{fig:relightining1}, \cref{fig:relightining2}, \cref{fig:relightining3}. For this experiment, we decouple the light direction from the camera to showcase the effectiveness of our method. By adjusting the spotlight angle, we control the light's coverage area. Additionally, modifying the roughness of the Gaussians results in changes to specularity: smoother surfaces exhibit greater reflectivity, while rougher surfaces reduce specularity. We strongly recommend reviewing the supplementary videos, which effectively demonstrate the relighting capabilities of our method.

\subsection{Additional Experiments on Simulating Body Movements}
Capturing detailed changes during body movements in static images is challenging; hence, we provide videos for visualization. We highly recommend reviewing supplementary videos, as they effectively demonstrate how our method enables realistic body movement simulations with minimal artifacts. To simulate body movements, we utilize GaMeS \cite{waczynska2024games} reparameterization combined with parametric or cage-based physical simulation.

\subsection{Additional Experiments on Decomposition}
 The supplementary videos include demonstration how our method separates material properties.

\subsection{Additional Results for Geometry}
\begin{figure*}[b!]
    \centering

    \includegraphics[width=0.65\textwidth, trim=0 0 0 0, clip]{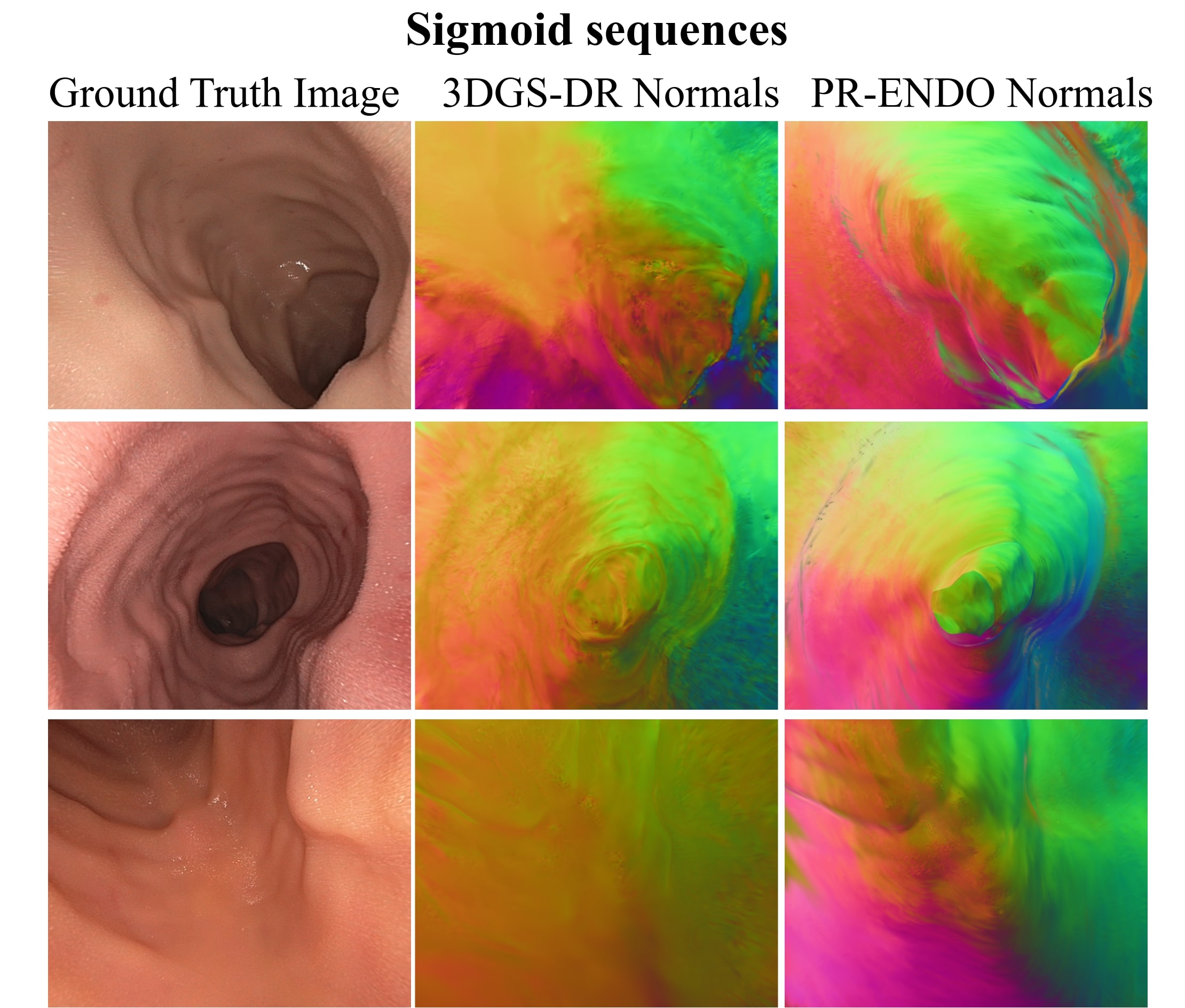}
    
    \caption{\textbf{Normal comparison across sigmoid C3VD scenes.} A single random capture is presented for each C3VD sequence. Our method generates higher-quality normals aligned with the ground truth geometry compared to the state-of-the-art 3DGS-DR \cite{ye2024gsdr}.}
    \label{fig:normals_1}
\end{figure*}

\begin{figure*}
    \centering

    \includegraphics[width=0.65\textwidth, trim=0 0 0 0, clip]{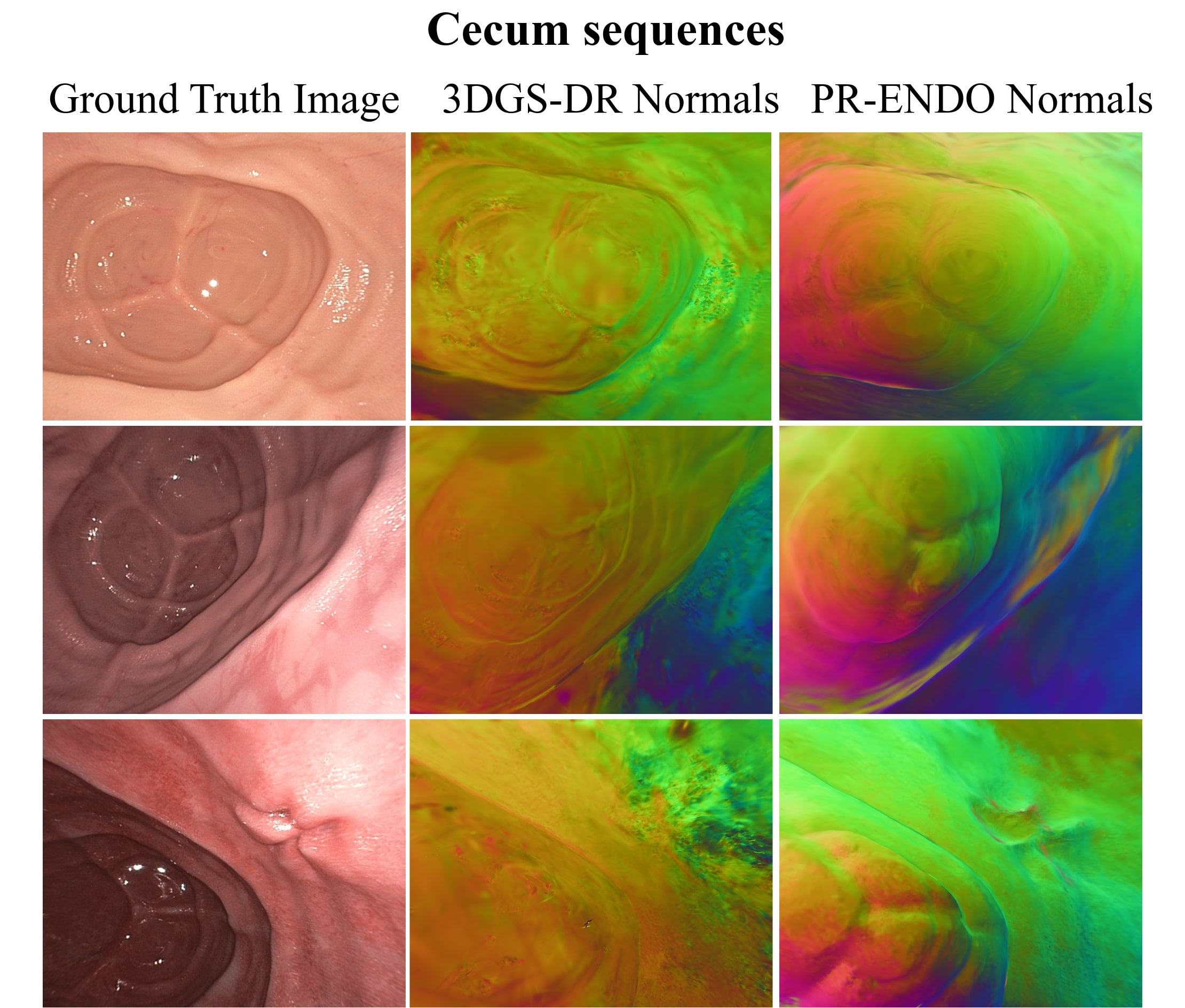}
    
    \caption{\textbf{Normal comparison across cecum C3VD scenes.} A single random capture is presented for each C3VD sequence. Our method generates higher-quality normals aligned with the ground truth geometry compared to the state-of-the-art 3DGS-DR \cite{ye2024gsdr}.}
    \label{fig:normals_2}
\end{figure*}

\begin{figure*}
    \centering

    \includegraphics[width=0.7\textwidth, trim=0 0 0 0, clip]{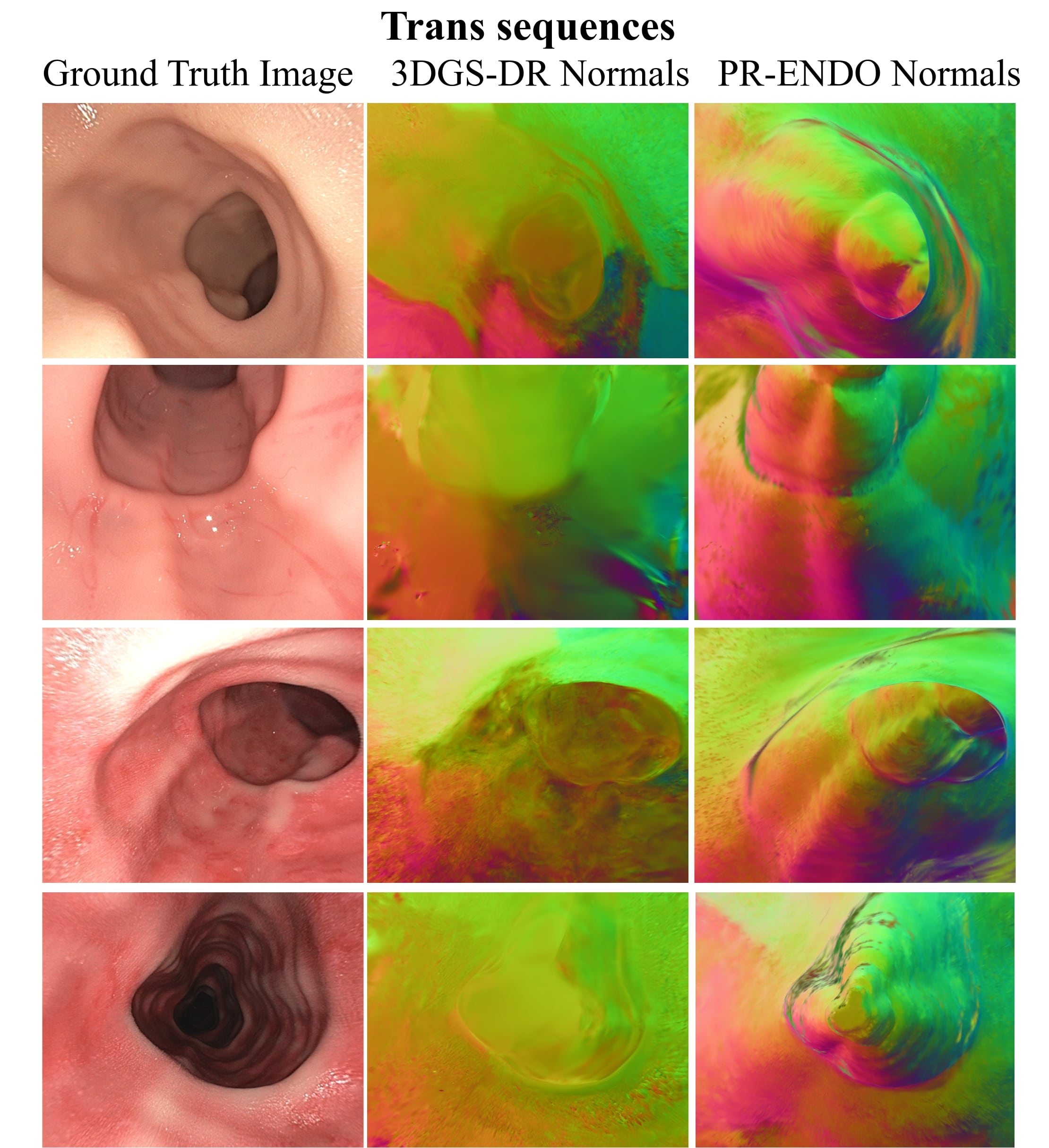}
    
    \caption{\textbf{Normal comparison across trans C3VD scenes.} A single random capture is presented for each C3VD sequence. Our method generates higher-quality normals aligned with the ground truth geometry compared to the state-of-the-art 3DGS-DR \cite{ye2024gsdr}.}
    \label{fig:normals_3}
\end{figure*}

In \cref{fig:normals_1}, \cref{fig:normals_2}, \cref{fig:normals_3}, we present an overview of the generated normals for each C3VD scene, alongside a comparison to normals produced by 3DGS-DR. The results demonstrate that our normals are more accurate and closely aligned with the ground truth geometry.

\end{document}